\documentclass[letterpaper]{article} 
\usepackage{aaai24}  
\usepackage{times}  
\usepackage{helvet}  
\usepackage{courier}  
\usepackage[hyphens]{url}  
\usepackage{graphicx} 
\usepackage{natbib}  
\usepackage{caption} 
\usepackage{algorithm}
\usepackage{algorithmic}

\usepackage{newfloat}
\usepackage{listings}
\DeclareCaptionStyle{ruled}{labelfont=normalfont,labelsep=colon,strut=off} 
\floatstyle{ruled}
\newfloat{listing}{tb}{lst}{}
\floatname{listing}{Listing}
\usepackage{amsmath,amsfonts}

\title{Multi-Source Collaborative Gradient Discrepancy Minimization for Federated Domain Generalization}
\author {
    Yikang Wei\textsuperscript{\rm 1,\rm 2},
    Yahong Han\textsuperscript{\rm 1,\rm 2}\thanks{Corresponding Author}
}
\affiliations {
    \textsuperscript{\rm 1}College of Intelligence and Computing, Tianjin University, Tianjin, China\\
    \textsuperscript{\rm 2}Tianjin Key Lab of Machine Learning, Tianjin University, Tianjin, China\\
    \{yikang,yahong\}@tju.edu.cn
}

\usepackage{bibentry}

\begin{document}

\maketitle

\begin{abstract}
Federated Domain Generalization aims to learn a domain-invariant model from multiple decentralized source domains for deployment on unseen target domain. Due to privacy concerns, the data from different source domains are kept isolated, which poses challenges in bridging the domain gap. To address this issue, we propose a Multi-source Collaborative Gradient Discrepancy Minimization (MCGDM) method for federated domain generalization. Specifically, we propose intra-domain gradient matching between the original images and augmented images to avoid overfitting the domain-specific information within isolated domains. Additionally, we propose inter-domain gradient matching with the collaboration of other domains, which can further reduce the domain shift across decentralized domains. Combining intra-domain and inter-domain gradient matching, our method enables the learned model to generalize well on unseen domains. Furthermore, our method can be extended to the federated domain adaptation task by fine-tuning the target model on the pseudo-labeled target domain. The extensive experiments on federated domain generalization and adaptation indicate that our method outperforms the state-of-the-art methods significantly.
\end{abstract}

\section{Introduction}

\begin{figure}[ht]
    \centering\includegraphics[width=0.49\textwidth]{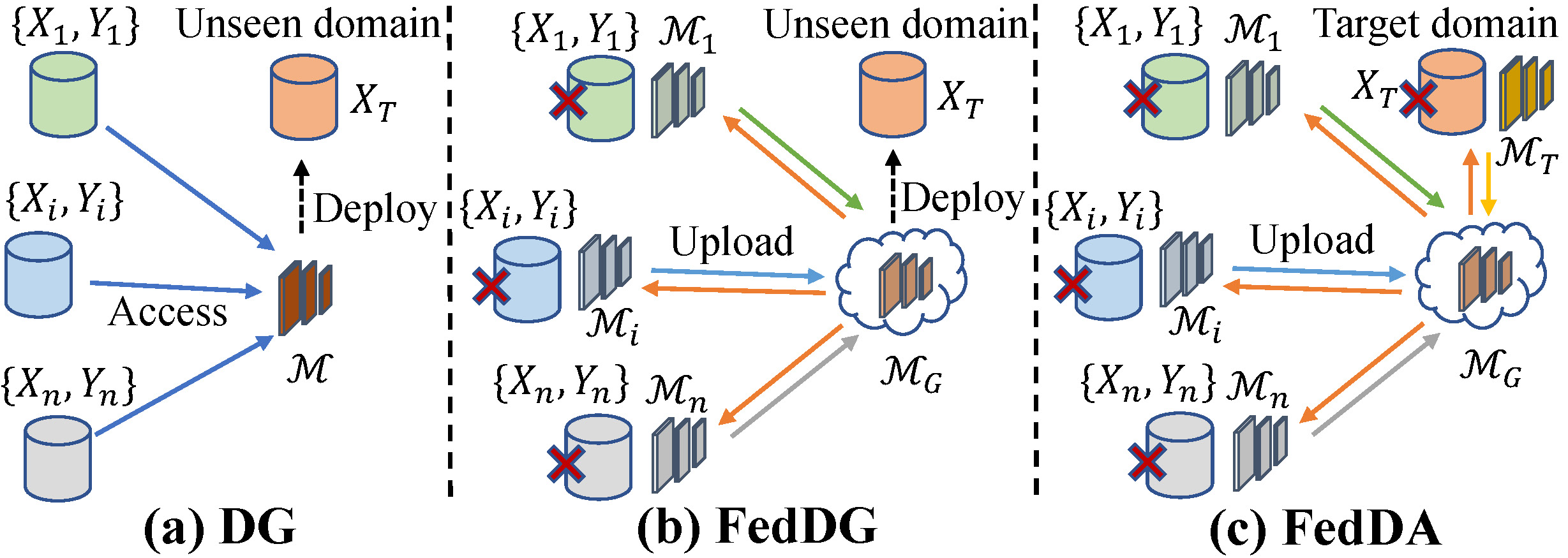}
    \centering\caption{(a) DG assumes that the data from multiple source domains $\{X_{i},Y_{i}\}_{i=1}^{n}$ can be accessed simultaneously to learn a generalized model $\mathcal{M}$ for deployment on the unseen domain $X_{T}$. (b) FedDG assumes that the data from different source domains are decentralized, but the local models $\{\mathcal{M}_{i}\}_{i=1}^{n}$ of different domains can be collaboratively trained and aggregated with a parameter server. (c) FedDA assumes that an additional unlabeled target domain $X_{T}$ can be accessed on server side for improving the performance.}
    \label{fig1}
\end{figure}

Domain generalization (DG) \cite{zhou2021domain,wang2022generalizing} aims to learn a domain-invariant model from multiple labeled source domains for deployment on the unseen target domains. The conventional DG methods \cite{wang2020learning,huang2020self,xu2021fourier} assume that the data from different source domains are centralized for conducting domain generalization, e.g. reducing domain shift across multiple source domains to learn domain-invariant model, as shown in Figure \ref{fig1}(a). Considering data privacy issues \cite{ye2023heterogeneous,huang2023generalizable,Huang_2023_CVPR,huang2023federated,zhou2023fedfa}, the data from different source domains are kept decentralized and only can be accessed on the isolated local clients. This data decentralization scenario brings great challenges to overcome the domain shift on the decentralized source domains.

To utilize the decentralized source domains, Federated Domain Generalization (FedDG) \cite{yuan2023collaborative,wu2021collaborative,liu2021feddg} collaboratively trains local models on the local clients and then aggregates the local models on the server side, as shown in Figure \ref{fig1}(b). Since the domain shift is partly caused by the discrepancy in image styles, existing FedDG methods synthesize images with novel styles by interpolating shared style information across domains \cite{liu2021feddg,chen2023federated} or training data generators within isolated domains \cite{xu2023federated} to mitigate the domain shift. However, when the target domain is unseen, the generated styles may not fully cover the characteristics of the unseen domain, resulting in limited generalization performance. Additionally, sharing style information across decentralized source domains raises privacy concerns, as it may risk privacy leakage.

Unlike existing FedDG methods aiming to expand the source domains, we explore the possibility of learning the domain-invariant model by reducing the domain shift on decentralized source domains without sharing data information. The domain shift for FedDG mainly comes from two aspects. Firstly, there exists domain gap between source domains and the unseen target domain, so the local models trained on source domains tend to be domain-specific and perform poorly on the unseen target domain. Secondly, the different source domains are isolated, so the domain shift between isolated source domains further hinders the generalization performance on unseen target domain.

For the above domain shift problems, this paper aims to learn the intrinsic semantic information within isolated source domains and reduce the domain shift across decentralized source domains for obtaining a domain-invariant model. Inspired by the hypothesis that gradient discrepancy between domains hinders domain-invariant learning \cite{mansilla2021domain,du2021cross}, where conflicting gradients across domains indicate that the model’s optimization direction becomes domain-specific, we propose a method called Multi-source Collaborative Gradient Discrepancy Minimization (MCGDM) for FedDG. Our method contains two key components. For avoiding the domain-specific information learned within the isolated source domains, we propose intra-domain gradient matching, which minimizes the gradient discrepancy between the original images and the augmented images to learn the intrinsic semantic information. Additionally, for bridging the domain gap across decentralized source domains, we propose inter-domain gradient matching, which minimizes the gradient discrepancy between the current model and the models of other domains. Combining the intra-domain and inter-domain gradient matching, the domain shift within isolated source domains and across decentralized source domains can be mitigated effectively so that a domain-invariant model can be learned. Furthermore, the proposed MCGDM method can be extended to the Federated Domain Adaptation (FedDA) task \cite{feng2021kd3a,wu2021collaborative,wei2022multi,10096982,wei2022dual,wu2023domain}, where the unlabeled target domain $X_{T}$ can be accessed on the server side for fine-tuning the target domain model, as illustrated in Figure \ref{fig1}(c). The extensive experiments on multiple datasets indicate the effectiveness of our method on FedDG and FedDA tasks.

The contributions of our work can be summarized as follows:

\begin{itemize}
  \item To avoid the domain-specific information learned by local models, we propose intra-domain gradient matching, which minimizes the gradient discrepancy between original images and augmented images for learning the intrinsic semantic information.
  \item To reduce the domain shift across decentralized source domains, we propose inter-domain gradient matching, which minimizes the gradient discrepancy between the current model and the models from other domains.
  \item By combining intra-domain gradient matching and inter-domain gradient matching, the domain shift can be reduced within isolated source domains and across decentralized source domains to generalize well on unseen target domain. Furthermore, our method can be expanded to the FedDA task. The experiments on FedDG and FedDA datasets indicate the effectiveness of our method.
\end{itemize}

\section{Related Work}

\subsection{Domain Generalization}
Conventional domain generalization utilizes multiple centralized source domains to learn a domain-invariant model for generalizing well on unseen target domain. The existing domain generalization methods can be categorized as (1) data augmentation methods, (2) domain-invariant representation learning methods, and (3) other learning strategies, such as self-supervised learning, and ensemble learning. The data augmentation methods aim to expand the source domains for improving the generalization performance of the model on unseen domain, such as L2A-OT \cite{zhou2020learning} and DDAIG \cite{zhou2020deep} train a data generator to synthesize out-of-distribution images, FACT \cite{xu2021fourier}, MixStyle \cite{zhou2021mixstyle}, and StyleNeo \cite{kang2022style} interpolate the styles between different domains to generate images or features with novel styles. The domain-invariant representation learning methods aim to learn the intrinsic semantic representations or reduce the domain shift across multiple source domains for generalizing well on the unseen target domain, such as RSC \cite{huang2020self} and CDG \cite{du2022cross} suppress the domain-specific representations to learn the intrinsic semantic representations across different domains. Furthermore, the other learning strategies such as self-supervised learning methods JiGen \cite{carlucci2019domain} and EISNet \cite{wang2020learning} utilize the jigsaw auxiliary task to learn the domain-invariant representations, the ensemble learning method DAEL \cite{zhou2021domain} distills the knowledge from the ensemble of other domains to learn domain-invariant model. Different from these methods which need to access the data from different source domains simultaneously, our method conducts multi-source collaborative federated domain generalization under the data decentralization scenario.

\begin{figure*}[htb]
    \centering\includegraphics[width=0.95\textwidth]{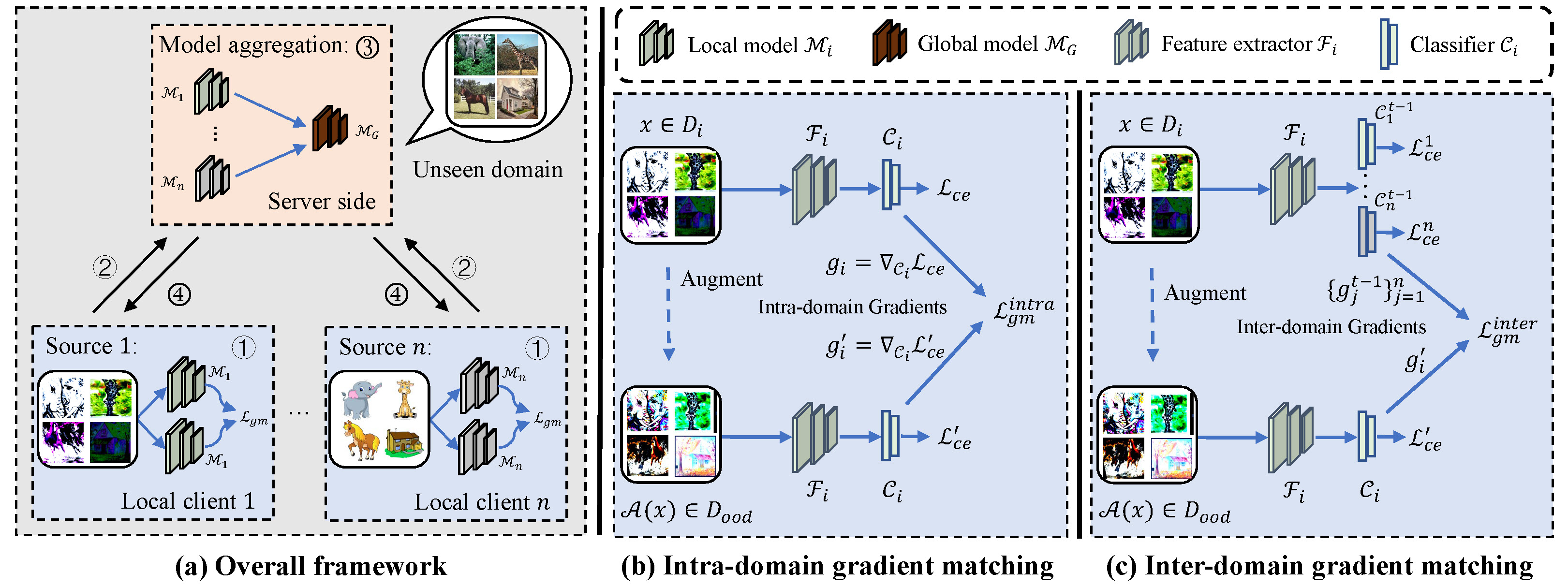}
    \centering\caption{(a) The overall framework of our method, where the local source domain models $\{\mathcal{M}_{i}\}_{i=1}^{n}$ are trained locally by conducting gradient matching and then are aggregated on the server side to obtain the generalizable global model $\mathcal{M}_{G}$. (b) The intra-domain gradient matching between the original images and the augmented images is conducted on local clients for learning the intrinsic semantic information within the domain. (c) The inter-domain gradient matching between the current classifier head $\mathcal{C}_{i}$ and the classifier heads from other domains $\{\mathcal{C}_{j}^{t-1}\}_{j=1}^{n}$ is conducted on the local clients for reducing the domain shift across decentralized source domains.}
    \label{fig2}
\end{figure*}

\subsection{Federated Domain Generalization}
Federated domain generalization learns a domain-invariant global model from multiple decentralized source domains for generalizing well on unseen target domain. The existing FedDG methods mainly focus on (1) data augmentation on decentralized source domains and (2) model aggregation on the server side. The data augmentation methods e.g. ELCFS \cite{liu2021feddg} and CCST \cite{chen2023federated} share the style information across decentralized source domains to generate the images with novel styles. The FADH \cite{xu2023federated} trains the data generators on the isolated source domains to synthesize images. And COPA \cite{wu2021collaborative} utilizes the classifier heads from other domains to learn the domain-invariant representations with the assistance of augmented images, e.g. RandAug \cite{cubuk2020randaugment}. These data augmentation methods focus on generating out-of-distribution images under the data decentralization scenario and conduct domain-invariant learning between the original images and augmented images. Different from these methods, our method focuses on learning domain-invariant models from the perspective of gradient discrepancy minimization and can be combined with the existing data augmentation methods. The other FedDG methods focus on the model aggregation stage on the server side for obtaining a fair model across different domains, which tends to be domain-invariant on the unseen domain. GA \cite{zhang2023federated} aggregates the local source models on the server side with different weights, and CASC \cite{yuan2023collaborative} aggregates the local source models with layer-wise weights. Different from these methods, we focus on learning the domain-invariant local models on decentralized source domains, which can achieve better generalization performance.

\section{Methodology}

\subsection{Overview}
For the domain generalization task, there are $n$ source domains $\{D_{i}\}_{i=1}^{n}$, each domain $D_{i}$ contains $N_{i}$ labeled images $\{x_{i},y_{i}\}_{i=1}^{N_{i}}$. A domain-invariant model is expected to be learned on source domains $\{D_{i}\}_{i=1}^{n}$ for deployment on the out-of-distribution unseen target domain $D_{t}$. There exists domain shift between source domains and unseen target domain. Here we consider the covariate shift, the marginal distribution $P(\mathcal{X})$ of the input images differs but the conditional label distribution $P(\mathcal{Y}|\mathcal{X})$ keeps the same across source domains and unseen target domain. With this assumption, a domain-invariant model learned on source domains can generalize well on unseen target domain by reducing the domain shift.

Under the data decentralization scenario, data from different source domains $\{D_{i}\}_{i=1}^{n}$ cannot be accessed simultaneously. In this work, we collaboratively train the decentralized source domains $\{D_{i}\}_{i=1}^{n}$ by the federated average \cite{mcmahan2017communication} to obtain a domain-invariant global model for applying on the unseen target domain $D_{t}$, as shown in Figure \ref{fig2}(a). Specifically, the local models $\{\mathcal{M}_{i}\}_{i=1}^{n}$ are trained on each isolated source domain $\{D_{i}\}_{i=1}^{n}$ and then are aggregated on the server side to obtain a global model $\mathcal{M}_{G}$. Each local model $\mathcal{M}_{i}$ contains a feature extractor $\mathcal{F}_{i}$ and a classifier head $\mathcal{C}_{i}$. There are 4 steps to collaboratively train the decentralized source domains $\{D_{i}\}_{i=1}^{n}$. In step 1, the local models $\{\mathcal{M}_{i}\}_{i=1}^{n}$ are trained on the isolated local clients via intra-domain and inter-domain gradient matching for learning intrinsic semantic information and reducing the domain shift across decentralized source domains. In step 2 and 3, the local models $\{\mathcal{M}_{i}\}_{i=1}^{n}$ are uploaded to the server side and are aggregated by parameter averaging to obtain a global model $\mathcal{M}_{G}$. In step 4, the global model $\mathcal{M}_{G}$ and classifier heads $\{\mathcal{C}_{i}^{t-1}\}_{i=1}^{n}$ of different domains are uploaded to local clients for the next round of training. After several rounds of training, a domain-invariant global model $\mathcal{M}_{G}$ can be learned across decentralized source domains and deployed on the unseen domain $D_{t}$.

\subsection{Intra-domain Gradient Matching}
As there are labeled source domain data on the isolated clients, we train the local model $\mathcal{F}_{i}\circ\mathcal{C}_{i}$ by cross-entropy loss as follows:

\begin{equation}
    \mathcal{L}_{ce}(x;\mathcal{F}_{i}\circ\mathcal{C}_{i})=-\sum_{i=1}^{B}y_{i}\log{\mathcal{F}_{i}\circ\mathcal{C}_{i}(x_{i})},
    \label{eqc1}
\end{equation}

where $B$ is the batch size of input images.

As mentioned above, it is challenging to learn the domain-invariant model on the local clients because the data from different source domains are decentralized. For example, the local model $\mathcal{M}_{i}$ trained only on the isolated client tends to be domain-specific for the source domain $D_{i}$. Inspired by the existing FedDG works \cite{wu2021collaborative,zhou2021domain}, which can learn the domain-invariant representations by enforcing the consistency between original images and augmented images on feature space or predictions, we learn the domain-invariant model with the assistance of augmented images to filter the domain-specific information on the isolated client. Different from the existing methods \cite{wu2021collaborative}, we enforce the consistency between original images and augmented images on parameter space to ensure local model updating toward the intrinsic semantic information, as shown in Figure \ref{fig2}(b). Specifically, we propose intra-domain gradient matching between original images and augmented images on the isolated source domains as follows:

\begin{equation}
    \mathcal{L}_{gm}^{intra}=1-sim(g_{i},g'_{i}).
    \label{eqc2}
\end{equation}

The $sim(\cdot,\cdot)$ is the cosine similarity. $g_{i}$ is the gradient of classifier $\mathcal{C}_{i}$ on original images:

\begin{equation}
    g_{i}=\nabla_{\mathcal{C}_{i}}\mathcal{L}_{ce}(x;\mathcal{F}_{i}\circ\mathcal{C}_{i}),
    \label{eqc3}
\end{equation}

where $\mathcal{L}_{ce}(x;\mathcal{F}_{i}\circ\mathcal{C}_{i})$ is the cross-entropy loss on original images.

And $g'_{i}$ is the gradient of classifier $\mathcal{C}_{i}$ on augmented images $\mathcal{A}_{x}$:

\begin{equation}
    g'_{i}=\nabla_{\mathcal{C}_{i}}\mathcal{L}_{ce}(\mathcal{A}(x);\mathcal{F}_{i}\circ\mathcal{C}_{i}).
    \label{eqc5}
\end{equation}

By using the intra-domain gradient matching between the original image $x$ and augmented image $\mathcal{A}(x)$, the local model tends to update toward the intrinsic semantic information contained on the original image $x$ and augmented image $\mathcal{A}(x)$. Following previous work \cite{wu2021collaborative,zhou2021domain}, the augmented images $\mathcal{A}(x)$ are generated by RandAugment (RandAug) \cite{cubuk2020randaugment}. Our method also can be combined with other style augmentation strategies, which is validated in the experiments section.

\subsection{Inter-domain Gradient Matching}

Due to the different source domains being isolated, it is challenging to bridge the domain gap across decentralized source domains. Different from the existing FedDG methods \cite{liu2021feddg,chen2023federated} sharing data information across domains, we use the classifier heads $\{\mathcal{C}_{j}^{t-1}\}_{j=1}^{n}$ from other domains as the bridges to reduce the domain shift, as shown in Figure \ref{fig2}(c). Specifically, we conduct inter-domain gradient matching for reducing the domain shift across decentralized source domains, where the gradient discrepancy between the current classifier head and the classifier heads from other domains are reduced:

\begin{equation}
    \mathcal{L}_{gm}^{inter}=\sum_{j=1}^{n}(1-sim(g'_{i},g_{j}^{t-1})).
    \label{eqc6}
\end{equation}

$g'_{i}$ is the gradient of current classifier $\mathcal{C}_{i}$ on augmented images $\mathcal{A}(x)$:

\begin{equation}
    g'_{i}=\nabla_{\mathcal{C}_{i}}\mathcal{L}_{ce}(\mathcal{A}(x);\mathcal{F}_{i}\circ\mathcal{C}_{i}),
    \label{eqc7}
\end{equation}

$g_{j}^{t-1}$ is the gradient of classifier $\mathcal{C}_{j}^{t-1}$ from $j$-th source domain on the original images:

\begin{equation}
    g_{j}^{t-1}=\nabla_{\mathcal{C}_{j}^{t-1}}\mathcal{L}_{ce}(x;\mathcal{F}_{i}\circ\mathcal{C}_{j}^{t-1}).
    \label{eqc8}
\end{equation}

By conducting inter-domain gradient matching between the current classifier head $\mathcal{C}_{i}$ and the classifier heads from different domains $\{\mathcal{C}_{j}^{t-1}\}_{j=1}^{n}$, the domain shift can be reduced across decentralized source domains.

On the local client, the overall loss function contains classification loss on the original images and augmented images, intra-domain gradient matching loss, and inter-domain gradient matching loss as follows:

\begin{equation}
\begin{split}
\mathcal{L}_{loc}=\frac{1}{2}(\mathcal{L}_{ce}(x;\mathcal{F}_{i}\circ\mathcal{C}_{i})+\mathcal{L}_{ce}(\mathcal{A}(x);\mathcal{F}_{i}\circ\mathcal{C}_{i})) \\
   +\lambda\mathcal{L}_{gm}^{intra} +(1-\lambda)\mathcal{L}_{gm}^{inter},
\end{split}
\label{eqc9}
\end{equation}


where the $\lambda$ is the hyper-parameter for balancing the intra-domain gradient matching loss $\mathcal{L}_{gm}^{intra}$ and inter-domain gradient matching loss $\mathcal{L}_{gm}^{inter}$.

\textbf{Remark:} Conducting gradient matching on overall model $\mathcal{F}_{i}\circ\mathcal{C}_{i}$ or feature extractor $\mathcal{F}_{i}$ will lead to large computational cost. So we only minimize the gradient discrepancy on classifier head $\mathcal{C}_{i}$, and the domain-invariant model can be learned by end-to-end training with the gradient matching loss. Furthermore, compared with sharing the data information of other domains, utilizing the classifier heads as the bridges have little risk of privacy leakage \cite{wu2021collaborative,jeong2021federated}.

\subsection{Model Aggregation}
As shown in Figure \ref{fig2}(a), after conducting local training e.g. 1 epoch on the isolated source domains, the local models $\{\mathcal{M}_{i}\}_{i=1}^{n}$ from different source domains are uploaded to server-side to conduct model aggregation in Step 3:

\begin{equation}\label{eqc10}
  \mathcal{M}_{G} = \sum_{i=1}^{n}\frac{N_{i}}{N_{total}}\mathcal{M}_{i},
\end{equation}

where the $N_{total}=\sum_{i=1}^{n}N_{i}$ is the sum of all samples. After conducting model aggregation on the server side, the global model $\mathcal{M}_{G}$ and the classifier heads $\{\mathcal{C}_{i}^{t-1}\}_{i=1}^{n}$ from different source domains are uploaded to local clients for the next round training. The global model $\mathcal{M}_{G}$ is used as the initial model for local training and the classifier heads $\{\mathcal{C}_{i}^{t-1}\}$ from different source domains are used to collaboratively bridge the domain gap across domains. We conduct $E$ rounds of local training on isolated source domains and model aggregation on the server side until the global model convergence.

\subsection{Extend to Federated Domain Adaptation}
Moreover, our method can be extended to the FedDA task. Different from FedDG where the target domain is unseen, FedDA \cite{feng2021kd3a,wei2022multi} assumes that unlabeled target domain data $X_{T}$ can be accessed on the server side for further improving the performance on the target domain, as shown in Figure \ref{fig1}(c). For utilizing the unlabeled target domain, the target domain model $\mathcal{M}_{T}$ is trained on the pseudo-labeled target domain data:

\begin{equation}
    \mathcal{L}_{ce}(x;\mathcal{F}_{T}\circ\mathcal{C}_{T})=-\sum_{i=1}^{B}\hat{y}_{T}\log{\mathcal{F}_{T}\circ\mathcal{C}_{T}(x_{T})},
    \label{eq_tar}
\end{equation}

where we utilize the domain-invariant local source models $\{\mathcal{M}_{i}\}_{i=1}^{n}$ learned by our MCGDM method to generate the high-confident pseudo-labels $\hat{y}_{T}$ on target domain following previous works \cite{feng2021kd3a}. Then the local source domain models $\{\mathcal{M}_{i}\}_{i=1}^{n}$ and target domain model $\mathcal{M}_{T}$ are aggregated to obtain the global model $\mathcal{M}_{G}$, which is used as the initial model for the next round of training on source domains and target domain. By fine-tuning the global model $\mathcal{M}_{G}$ on the pseudo-labeled target domain, the obtained target domain model $\mathcal{M}_{T}$ can achieve better performance on the target domain.

\section{Experiments}

\subsection{Datasets}
\textbf{Federated Domain Generalization Datasets.} We conduct FedDG experiments on three image classification datasets including \textbf{PACS}\cite{li2017deeper}, \textbf{VLCS}\cite{ghifary2015domain}, and \textbf{Office-Home}\cite{venkateswara2017deep}. \textbf{PACS} contains 9,991 images of 7 categories from four domains: Art-painting (A), Cartoon (C), Photo (P), and Sketch (S), the data of each domain are split into 80\% for training and 20\% for testing. \textbf{Office-Home} contains 15,500 images of 65 categories from four domains: Artist (A), Clipart (C), Product (P), and Real-world (R), each domain is split into 90\% as the training set and 10\% as the test set. \textbf{VLCS} contains 10,729 images of 5 categories from four domains: Pascal (P), LabelMe (L), Caltech (C), and Sun (S), each domain is split into 80\% for training and 20\% for testing. The leave-one-domain-out protocol \cite{zhou2021mixstyle} is used to evaluate the generalization performance on one domain and train the model on the rest source domains.

\textbf{Federated Domain Adaptation Datasets.} We conduct FedDA experiments on two image classification datasets including \textbf{Digit-5} and \textbf{Office-Caltech10}. \textbf{Digit-5} contains five domains MNIST(mt), MNIST-M(mm), SVHN(sv), Syn(syn), and USPS(up) of 10 categories. \textbf{Office-Caltech10} contains four domains Amazon(A), Caltech(C), Webcam(W), and Dslr(D) of 10 categories. Following the setting of previous works \cite{feng2021kd3a,wu2021collaborative}, the model is trained on the labeled source domains and an unlabeled target domain, and the results on the target domain are reported.

\subsection{Implementation Details}

Following the previous works \cite{yuan2023collaborative,wu2021collaborative}, we use the ResNet-18 \cite{he2016deep} pre-trained on ImageNet as the backbone for PACS, Office-Home, and VLCS datasets. The SGD optimizer with momentum 0.9 and weight decay 5e-4 is adopted for PACS, Office-Home, and VLCS. For PACS and VLCS, the batch size is 16 and the initial learning rate is 0.001, decayed by the cosine schedule from 0.001 to 0.0001 during training. For Office-Home, the batch size is 30 and the initial learning rate is 0.002, decayed by cosine scheduled from 0.002 to 0.0001. The hyper-parameter $\lambda$ on Equation \ref{eqc9} is 0.3 for PACS, 0.8 for Office-Home and VLCS.

For each local client, we train the local model 1 epoch and then upload the local model to the server side for conducting model aggregation. The local training on the client side and global model aggregation on the server side are conducted iteratively. The total training rounds $E$ is 40 on PACS, VLCS, and Office-Home datasets. All experiments are repeated three times with different random seeds and the mean accuracy (\%) is reported.

For the federated domain adaptation datasets, we use the three-layer CNN as the backbone for Digit-5 and the ResNet-101 pre-trained on ImagesNet as the backbone for Office-Caltech10 following the setting of previous work KD3A \cite{feng2021kd3a}. The total training rounds are 50 for Digit-5 and 40 for Office-Caltech10. Following previous work \cite{feng2021kd3a}, we generate the pseudo-labels on the unlabeled target domain by knowledge vote. We implement our method with PyTorch and use a single NVIDIA RTX3090. Moreover, we reproduce the results on PACS dataset with MindSpore framework \cite{mindspore}.

\begin{table}[htb]
\centering
\small{
  \setlength{\tabcolsep}{1.3mm}{
  \begin{tabular}{lccccc}
  \hline
  Methods & A & C & P & S & Avg \\
  \hline
  DeepAll \cite{zhou2021domain}   & 77.0 & 75.9 & 96.0 & 69.2 & 79.5 \\
  JiGen \cite{carlucci2019domain} & 79.4 & 75.3 & 96.0 & 71.4 & 80.5 \\
  EISNet \cite{wang2020learning}  & 81.9 & 76.4 & 95.9 & 74.3 & 82.2 \\
  L2A-OT \cite{zhou2020learning}  & 83.3 & 78.2 & \textbf{96.2} & 73.6 & 82.8 \\
  DDAIG \cite{zhou2020deep}       & 84.2 & 78.1 & 95.3 & 74.7 & 83.1 \\
  FACT \cite{xu2021fourier}       & 85.4 & 78.4 & 95.2 & 79.2 & 84.5 \\
  MixStyle \cite{zhou2021mixstyle} & 84.1 & 78.8 & \underline{96.1} & 75.9 & 83.7 \\
  StyleNeo \cite{kang2022style}   & 84.4 & 79.2 & 94.9 & 83.2 & 85.4 \\
  RSC \cite{huang2020self}        & 83.4 & 80.3 & 96.0 & 80.9 & 85.2 \\
  CDG \cite{du2022cross}          & 83.5 & 80.1 & 95.6 & \underline{83.8} & \underline{85.8} \\
  DAEL \cite{zhou2021domain}      & 84.6 & 74.4 & 95.6 & 78.9 & 83.4 \\
  \hline
  FedAvg \cite{mcmahan2017communication} & 82.2 & 77.3 & 94.5 & 71.2 & 81.3 \\
  CASC \cite{yuan2023collaborative} & 82.0 & 76.4 & 95.2 & 81.6 & 83.8 \\
  GA \cite{zhang2023federated} & 83.2 & 76.9 & 94.0 & 82.9 & 84.3 \\
  ELCFS \cite{liu2021feddg} & 82.3 & 74.7 & 93.3 & 82.7 & 83.2 \\
  FADH \cite{xu2023federated} & 83.8 & 77.2 & 94.4 & \textbf{84.4} & 85.0 \\
  COPA \cite{wu2021collaborative} & 83.3 & 79.8 & 94.6 & 82.5 & 85.1 \\
  MCGDM (ours) & \textbf{86.2} & \underline{81.2} & 96.0 & 81.2 & \textbf{86.2} \\
  MCGDM\dag (ours) & \underline{86.1} & \textbf{81.4} & 96.0 & 81.2 & \textbf{86.2} \\
  \hline
  \end{tabular}
}
}
\caption{Accuracy(\%) on PACS dataset. We have bolded the best results and underlined the second results. \dag\ indicates the results reproduced with MindSpore framework.}
\label{pacs}
\end{table}

\begin{table}[htb]
\centering
\small{
  \setlength{\tabcolsep}{1.3mm}{
  \begin{tabular}{lccccc}
  \hline
  Methods & A & C & P &  R  & Avg \\
  \hline
  DeepAll \cite{zhou2021domain}   & 57.9 & 52.7 & 73.5 & 74.8 & 64.7 \\
  JiGen \cite{carlucci2019domain} & 53.0 & 47.5 & 71.5 & 72.8 & 61.2 \\
  EISNet \cite{wang2020learning}  & 56.8 & 53.3 & 72.3 & 73.5 & 64.0 \\
  L2A-OT \cite{zhou2020learning}  & \textbf{60.6} & 50.1 & 74.8 & \textbf{77.0} & 65.6 \\
  DDAIG \cite{zhou2020deep}       & 59.2 & 52.3 & 74.6 & \underline{76.0} & 65.5 \\
  MixStyle \cite{zhou2021mixstyle} & 58.7 & 53.4 & 74.2 & 75.9 & 65.5 \\
  StyleNeo \cite{kang2022style}   & 59.6 & 55.0 & 73.6 & 75.5 & 65.9 \\
  RSC \cite{huang2020self}        & 58.4 & 47.9 & 71.6 & 74.5 & 63.1 \\
  CDG \cite{du2022cross}          & 59.2 & 54.3 & \underline{74.9} & 75.7 & 66.0 \\
  DAEL \cite{zhou2021domain}      & 59.4 & 55.1 & 74.0 & 75.7 & \underline{66.1} \\
  \hline
  FedAvg \cite{mcmahan2017communication} & 58.2 & 51.6 & 73.1 & 73.8 & 64.2 \\
  GA \cite{zhang2023federated} & 58.8 & 54.3 & 73.7 & 74.7 & 65.4 \\
  ELCFS \cite{liu2021feddg} & 57.8 & 54.9 & 71.1 & 73.1 & 64.2 \\
  CCST \cite{chen2023federated} & 59.1 & 50.1 & 73.0 & 71.7 & 63.6 \\
  FADH \cite{xu2023federated} & \underline{59.9} & \underline{55.8} & 73.5 & 74.9 & 66.0 \\
  COPA \cite{wu2021collaborative} & 59.4 & 55.1 & 74.8 & 75.0 & \underline{66.1} \\
  MCGDM (ours) & \underline{59.9} & \textbf{56.7} & \textbf{75.3} & 75.2 & \textbf{66.8} \\
  \hline
  \end{tabular}
  }
}
\caption{Accuracy(\%) on Office-Home dataset. We have bolded the best results and underlined the second results.}
\label{office_home}
\end{table}

\begin{table}[htb]
\centering
\small{
  \setlength{\tabcolsep}{1.3mm}{
  \begin{tabular}{lccccc}
  \hline
  Methods & P & L & C & S & Avg \\
  \hline
  DeepAll \cite{zhou2021domain}   & 71.4 & 59.8 & 97.5 & 69.0 & 74.4 \\
  JiGen \cite{carlucci2019domain} & 74.0 & 61.9 & 97.4 & 66.9 & 75.1 \\
  L2A-OT \cite{zhou2020learning}  & 72.8 & 59.8 & \underline{98.0} & 70.9 & 75.4 \\
  RSC \cite{huang2020self}        & \underline{75.3} & 59.8 & 97.0 & 71.5 & 75.9 \\
  MixStyle \cite{zhou2021mixstyle} & 72.6 & 58.5 & 97.7 & 73.3 & 75.5 \\
  \hline
  FedAvg \cite{mcmahan2017communication} & 72.0 & 63.3 & 96.5 & 72.4 & 76.0 \\
  CASC \cite{yuan2023collaborative} & 72.0 & \underline{63.5} & 97.2 & 72.1 & \underline{76.2} \\
  ELCFS \cite{liu2021feddg} & 71.1 & 59.5 & 96.6 & \textbf{74.0} & 75.3 \\
  COPA \cite{wu2021collaborative} & 71.5 & 61.0 & 93.8 & 71.7 & 74.5 \\
  MCGDM (ours) & \textbf{75.6} & \textbf{64.3} & \textbf{98.6} & \underline{73.8} & \textbf{78.1} \\
  \hline
  \end{tabular}
  }
}
\caption{Accuracy(\%) on VLCS dataset. We have bolded the best results and underlined the second results.}
\label{vlcs}
\end{table}

\subsection{Comparison with State-of-the-Art Methods}

\textbf{Results on PACS.} PACS is a dataset with a large domain gap between different domains, e.g. different image styles. As shown in Table \ref{pacs}, our method achieves the best average accuracy of 86.2\% compared with the conventional DG methods and FedDG methods. Compared with the data augmentation FedDG methods, e.g. ELCFS, FADH, and COPA, our method focuses on learning the domain-invariant model across original images and augmented images, which achieves the 1\%-3\% improvement on average accuracy. Compared with the FedDG methods which calibrate the aggregation weights on the server side, e.g. GA and CASC, our method reduces the domain shift on local clients and achieves a 2\% improvement on average accuracy. Furthermore, our method even outperforms the conventional DG methods which can access the data from different source domains simultaneously, e.g. data augmentation methods FACT and domain-invariant representation learning methods RSC and CDG.

\textbf{Results on Office-Home.} As shown in Table \ref{office_home}, our method also achieves the best average accuracy of 66.8\% compared with other FedDG methods and conventional DG methods. Due to the domain gap between isolated source domains being ignored, the existing data style augmentation methods e.g. ELCFS and CCST cannot achieve significant improvements on average accuracy. Instead, our method reduces the domain shift across isolated source domains and achieves significant improvements. Compared with COPA, which fixes the classifier heads from other domains and learns the domain-invariant feature extractor, our method achieves better performance by conducting intra-domain and inter-domain gradient matching on local clients. We also conduct ablation study about different alignment strategies in the Ablation Study section to indicate the effectiveness of our gradient matching strategy.

\textbf{Results on VLCS.} We also conduct experiments on the VLCS dataset. As shown in Table \ref{vlcs}, our method achieves consistent improvement on Pascal, LabelMe, Caltech, and Sun domains. Compared with other FedDG methods, our method improves the average accuracy by 2\%-3\%. Compared with the state-of-the-art conventional DG methods, such as RSC and MixStyle, our method also achieves significant improvements with the privacy constraints. The results on VLCS indicate the effectiveness of our method to learn the domain-invariant local models for better generalization performance.

\subsection{Ablation Study}

\begin{table}
\centering
\small{
  \setlength{\tabcolsep}{0.9mm}{
  \begin{tabular}{ccccccccc}
    \hline
    Baseline & RandAug & $\mathcal{L}_{gm}^{intra}$ & $\mathcal{L}_{gm}^{inter}$ & A & C & P & S & Avg\\
    \hline
    \checkmark &  &  &  & 82.2 & 77.3 & 94.5 & 71.2 & 81.3  \\
    \checkmark & \checkmark &  &  & 85.2 & 78.6 & \textbf{96.0} & 75.8 & 83.9  \\
    \checkmark & \checkmark & \checkmark &  & 85.8 & 79.8 & 95.4 & 80.8 & 85.5  \\
    \checkmark & \checkmark &  & \checkmark & 85.6 & 79.8 & 94.8 & \textbf{81.3} & 85.4  \\
    \checkmark & \checkmark & \checkmark & \checkmark & \textbf{86.2} & \textbf{81.2} & \textbf{96.0} & 81.2 & \textbf{86.2}  \\
  \hline
\end{tabular}
}
}
\caption{Accuracy(\%) of each component on PACS dataset, combined with the data augmentation strategy RandAug.}
\label{tab_contributions_pacs_randaug}
\end{table}

\subsubsection{Contributions of Different Components}
As shown in Table \ref{tab_contributions_pacs_randaug}, the Baseline indicates training the decentralized source domains by FedAvg \cite{mcmahan2017communication} combined with MixStyle \cite{zhou2021mixstyle}, which interpolates the features within mini-batch. By utilizing the data augmentation strategy RandAug \cite{cubuk2020randaugment} to expand source domains, the performance on unseen target domain can be further improved. Our method aims to reduce the domain shift so that the learned model can generalize well on unseen domain. By combining the gradient matching loss $\mathcal{L}_{gm}^{intra}$ and $\mathcal{L}_{gm}^{inter}$ proposed in this paper, the best average accuracy of 86.2\% can be achieved.

\begin{figure}[htb]
    \centering\includegraphics[width=0.45\textwidth]{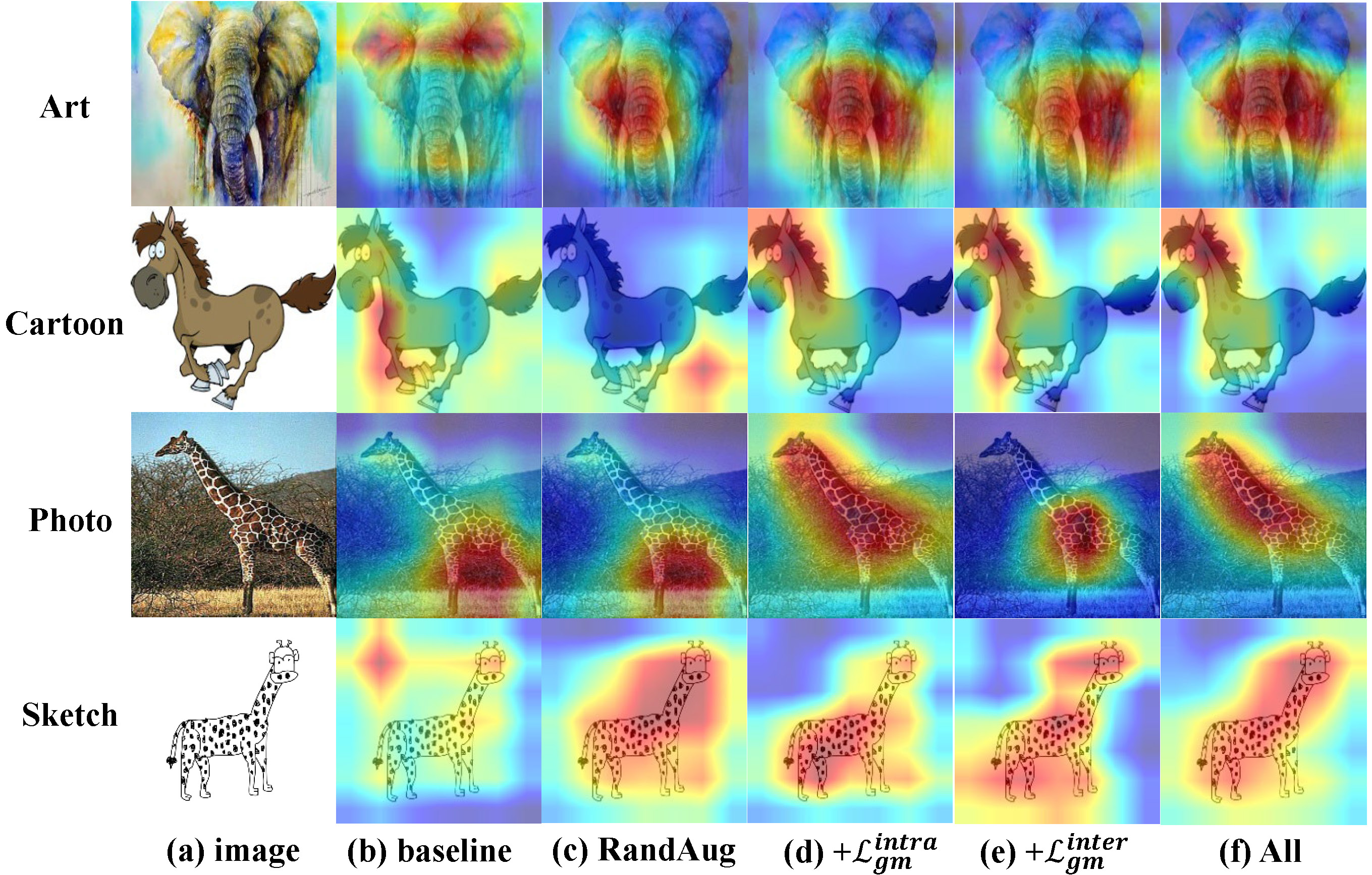}
    \centering\caption{Visualization by Grad-CAM on unseen domain.}
    \label{cam}
\end{figure}

Furthermore, we visualize the activation maps by Grad-CAM \cite{selvaraju2017grad}, as shown in Figure \ref{cam}. Combining with the $\mathcal{L}_{gm}^{intra}$ and $\mathcal{L}_{gm}^{inter}$, the learned model can focus on the most discriminative regions on the images from the unseen domain, which will lead to better generalizable performance on the unseen domain.

\begin{table}[htb]
\centering
\small{
\setlength{\tabcolsep}{1.5mm}{
\begin{tabular}{l c c c c c}
\hline
Method & A & C & P & S & Avg \\
\hline
Baseline & 85.2 & 78.6 & 96.0 & 75.8 & 83.9  \\
+ KL \cite{kang2022style} & 85.8 & 78.7 & 96.0 & 78.6 & 84.8  \\
+ SupCon \cite{khosla2020supervised} & 86.1 & 78.5 & \textbf{96.2} & 78.3 & 84.8  \\
+ CoFC \cite{wu2021collaborative} & 85.4 & 79.1 & 96.0 & 76.3 & 84.2  \\
+ MCGDM (ours) & \textbf{86.2} & \textbf{81.2} & 96.0 & \textbf{81.2} & \textbf{86.2} \\
\hline
\end{tabular}
}
}
\caption{Accuracy (\%) of different alignment strategies on PACS dataset.}
\label{align_pacs}
\end{table}

\subsubsection{Comparison with Different Alignment Strategies}
We also make a comparison with different alignment strategies between the original images and augmented images: (1) reducing the Kullback-Leibler (KL) divergence on logits-level between original images and augmented images, which is used by StyleNeo \cite{kang2022style}, (2) conducting Supervised Contrastive loss (SupCon) \cite{khosla2020supervised} on feature-level between the original images and augmented images, (3) Collaboration of Frozen Classifiers (CoFC) used by COPA \cite{wu2021collaborative}, which freezes the multiple classifier heads of other domains to learn the domain-invariant feature extractor. Different from these methods which align the features or logits between original images and augmented images, our method reduces the gradient discrepancy within isolated domains and across decentralized source domains. As shown in Table \ref{align_pacs}, our method outperforms other alignment strategies by about 2\% on average accuracy and achieves large improvement on hard domains, e.g. Art, Cartoon, and Sketch.

\subsubsection{Combined with Other Augmentation Strategy}
We also combine our gradient matching method with the data augmentation strategy Amplitude Mix (AM), which is the main component used by ELCFS and FACT to interpolate the style information across domains. Following previous work \cite{zhang2023federated}, we interpolate the amplitudes of different images within mini-batch to expand source domains. As shown in Table \ref{tab_contributions_pacs_am}, our method can improve the average accuracy by about 2\% by combining with AM on PACS dataset, which can indicate the effectiveness of our method in reducing the domain shift. In this work, we use the RandAug as the default data augmentation strategy.

\begin{table}
\centering
\small{
  \setlength{\tabcolsep}{1.3mm}{
  \begin{tabular}{ccccccccc}
    \hline
    Baseline & AM & $\mathcal{L}_{gm}^{intra}$ & $\mathcal{L}_{gm}^{inter}$ & A & C & P & S & Avg\\
    \hline
    \checkmark &  &  &  & 82.2 & 77.3 & \textbf{94.5} & 71.2 & 81.3  \\
    \checkmark & \checkmark &  &  & 83.3 & 77.1 & 93.3 & 76.8 & 82.6  \\
    \checkmark & \checkmark & \checkmark &  & 83.3 & 78.6 & 93.1 & 81.0 & 84.0  \\
    \checkmark & \checkmark &  & \checkmark & 83.6 & 78.3 & 93.1 & 80.6 & 83.9  \\
    \checkmark & \checkmark & \checkmark & \checkmark & \textbf{84.1} & \textbf{79.0} & 93.3 & \textbf{81.9} & \textbf{84.6}  \\
  \hline
\end{tabular}
}
}
\caption{Accuracy(\%) of each component on PACS dataset, combined with other data augmentation strategy.}
\label{tab_contributions_pacs_am}
\end{table}

\subsubsection{Experiments on Federated Domain Adaptation}
We make a comparison with the multi-source domain adaptation methods: (1) conventional unsupervised multi-source domain adaptation methods DCTN \cite{xu2018deep} and M3SDA \cite{peng2019moment}, where the source domains and target domain are centralized, (2) Source-free methods SHOT \cite{liang2020we}, DECISION \cite{ahmed2021unsupervised}, and CAiDA \cite{dong2021confident}, where only the pre-trained source domain models can be accessed on unlabeled target domain for adaptation, (3) FedDA methods FADA \cite{peng2019federated}, KD3A \cite{feng2021kd3a}, and COPA \cite{wu2021collaborative}, where the decentralized source domains and target domain are trained collaboratively.

\begin{table}[htbp]
\centering
\centering
\small{
\setlength{\tabcolsep}{0.4mm}{
\begin{tabular}{l c c c c c c}
\hline
Method & mt & mm & sv & syn & up & Avg \\
\hline
Source-only & 92.3 & 63.7 & 71.5 & 83.4 & 90.7 & 80.3  \\
DCTN \cite{xu2018deep} & 96.2 & 70.5 & 77.6 & 86.8 & 92.8 & 84.8  \\
M3SDA \cite{peng2019moment} & 98.4 & 72.8 & 81.3 & 89.6 & 96.2 & 87.7  \\
\hline
SHOT \cite{liang2020we} & 98.2 & 80.2 & 84.5 & 91.1 & 97.1 & 90.2  \\
DECISION \cite{ahmed2021unsupervised} & 99.2 & 93.0 & 82.6 & 97.5 & 97.8 & 94.0  \\
CAiDA \cite{dong2021confident} & 99.1 & \underline{93.7} & 83.3 & \textbf{98.1} & \underline{98.6} & 94.6  \\
\hline
FADA \cite{peng2019federated} & 91.4 & 62.5 & 50.5 & 71.8 & 91.7 & 73.6  \\
KD3A \cite{feng2021kd3a} & 99.2 & 87.3 & 85.6 & 89.4 & 98.5 & 92.0  \\
COPA \cite{wu2021collaborative} & \underline{99.4} & 89.8 & \underline{91.0} & 97.5 & \textbf{99.2} & \underline{95.4}  \\
MCGDM (ours) & \textbf{99.5} & \textbf{96.9} & \textbf{92.3} & \underline{97.8} & \textbf{99.2} & \textbf{97.1}  \\
\hline
\end{tabular}
}
}
\caption{Accuracy(\%) of different multi-source domain adaptation methods on Digit-5 dataset.}
\label{digit5}
\end{table}

\begin{table}[htbp]
\centering
\centering
\small{
\setlength{\tabcolsep}{1.0mm}{
\begin{tabular}{l c c c c c}
\hline
Method & A & C & D & W & Avg \\
\hline
Source-only & 86.1 & 87.8 & 98.3 & 99.0 & 92.8  \\
DCTN \cite{xu2018deep} & 92.7 & 90.2 & 99.0 & 99.4 & 95.3  \\
M3SDA \cite{peng2019moment} & 94.5 & 92.2 & 99.2 & 99.5 & 96.4  \\
\hline
SHOT \cite{liang2020we} & 96.4 & 96.2 & 98.5 & 99.7 & 97.7  \\
DECISION \cite{ahmed2021unsupervised} & 95.9 & 95.9 & \textbf{100} & 99.6 & 98.0  \\
CAiDA \cite{dong2021confident} & \underline{96.8} & \textbf{97.1} & \textbf{100} & \underline{99.8} & \underline{98.4}  \\
\hline
FADA \cite{peng2019federated} & 84.2 & 88.7 & 87.1 & 88.1 & 87.1  \\
KD3A \cite{feng2021kd3a} & \textbf{97.4} & 96.4 & 98.4 & 99.7 & 97.9  \\
COPA \cite{wu2021collaborative} & 95.8 & 94.6 & \underline{99.6} & \underline{99.8} & 97.5  \\
MCGDM (ours) & \textbf{97.4} & \underline{96.9} & \textbf{100} & \textbf{100} & \textbf{98.6}  \\
\hline
\end{tabular}
}
}
\caption{Accuracy(\%) of different multi-source domain adaptation methods on Office-Caltech10 dataset.}
\label{office_caltech10}
\end{table}

As shown in Table \ref{digit5} and Table \ref{office_caltech10}, our method outperforms other methods significantly. Different from the source-free methods, e.g., SHOT, DECISION, and CAiDA, our method can collaboratively train the isolated source domains and target domain and lead to better performance. Different from the FedDA methods e.g. FADA, KD3A, and COPA, our method conducts gradient matching on the local clients to reduce the domain shift, so that the learned local models on source domains can be generalized well on the target domain. By further fine-tuning the target model on the pseudo-labeled target domain, our method can achieve state-of-the-art performance and outperform other FedDA methods significantly.

\section{Conclusion}
In this paper, we propose a multi-source collaborative gradient discrepancy minimization method for federated domain generalization. Our main idea is to reduce the domain shift within isolated source domains and across decentralized source domains for learning the domain-invariant model, which can generalize on unseen target domain. Moreover, our method can be combined with the existing data augmentation strategies and improve the generalization performance on unseen target domain. Furthermore, we extend our method to the federated domain adaptation task by fine-tuning the target model on the pseudo-labeled target domain. The extensive experiments on federated domain generalization and adaptation datasets can validate the effectiveness of our method.

\section{Acknowledgments}
This work is supported by the National Natural Science Foundation of China (under Grant 62376186, 61932009) and the CAAI-Huawei MindSpore Open Fund.

\bibliographystyle{aaai24}
\bibliography{samples}

\end{document}